\documentclass[10pt,twocolumn,letterpaper]{article}

\usepackage[pagenumbers]{cvpr} 

\usepackage{graphicx}
\usepackage[outdir=latex/fig/]{epstopdf}
\usepackage{amsmath}
\usepackage{amssymb}
\usepackage{booktabs}
\usepackage{multirow}
\usepackage{comment}
\usepackage{algorithmic}
\usepackage[ruled,vlined]{algorithm2e}
\usepackage{balance}

\usepackage{caption}
\usepackage{subcaption}

\newtheorem{myDef}{\textsc{Definition}}

\newtheorem{mylem}{\textsc{Lemma}}

\newcommand{\Hquad}{\hspace{0.5em}} 

\usepackage[pagebackref,breaklinks,colorlinks]{hyperref}

\usepackage[capitalize]{cleveref}
\crefname{section}{Sec.}{Secs.}
\Crefname{section}{Section}{Sections}
\Crefname{table}{Table}{Tables}
\crefname{table}{Tab.}{Tabs.}

\def\cvprPaperID{3549} 
\def\confName{CVPR}
\def\confYear{2022}

\begin{document}

\title{Primitive3D: 3D Object Dataset Synthesis from Randomly Assembled Primitives}

\author{Xinke Li$^{1,}$\footnotemark[1]\quad Henghui Ding$^{2,3,}$\footnotemark[1] \quad Zekun Tong$^{1}$\quad Yuwei Wu$^{1}$\quad Yeow Meng Chee$^{1}$ \\
$^{1}$National University of Singapore \quad  $^{2}$ByteDance \quad $^{3}$ETH Zürich\\
\small{\texttt{\{xinke.li, zekuntong\}@u.nus.edu}} \quad \small{\texttt{henghui.ding@vision.ee.ethz.ch}} \quad \small{\texttt{\{wyw, ymchee\}@nus.edu.sg}}\\
}
\maketitle
\renewcommand{\thefootnote}{\fnsymbol{footnote}}
\footnotetext[1]{Equal contribution, Henghui was Xinke's internship mentor.}

\begin{abstract}

Numerous advancements in deep learning can be attributed to the access to large-scale and well-annotated datasets. However, such a dataset is prohibitively expensive in 3D computer vision due to the substantial collection cost. To alleviate this issue, we propose a cost-effective method for automatically generating a large amount of 3D objects with annotations. 
In particular, we synthesize objects simply by assembling multiple random primitives. 
These objects are thus auto-annotated with part labels originating from primitives. 
This allows us to perform multi-task learning by combining the supervised segmentation with unsupervised reconstruction. Considering the large overhead of learning on the generated dataset, we further propose a dataset distillation strategy to remove redundant samples regarding a target dataset. We conduct extensive experiments for the downstream tasks of 3D object classification. 
The results indicate that our dataset, together with multi-task pretraining on its annotations, achieves the best performance compared to other commonly used datasets. Further study suggests that our strategy can improve the model performance by pretraining and fine-tuning scheme, especially for the dataset with a small scale. In addition, pretraining with the proposed dataset distillation method can save 86\% of the pretraining time with negligible performance degradation. We expect that our attempt provides a new data-centric perspective for training 3D deep models.

\end{abstract}

\section{Introduction}
\label{sec:intro}

Deep learning has been shown to surpass prior state-of-the-art machine learning techniques in applications of 2D computer vision in the past several years~\cite{ding2018context,ding2019boundary,ding2020semantic}.
Such success is often attributed to the ease of acquiring large-scale, richly-annotated and diverse 2D image datasets~\cite{zhang2021prototypical}, \eg, ImageNet~\cite{deng2009imagenet} and COCO\cite{lin2014microsoft}. In the field of 3D, a broad variety of applications, such as self-driving vehicles~\cite{chen20153d}, augmented reality~\cite{gordon2006and}, and urban construction~\cite{chen2018performance}, can benefit from advancing 3D object understanding, which has long awaited high-quality datasets. Indeed, unlike the 2D counterparts, existing 3D object datasets are often limited in scale or lack of variety in annotation and instance 
diversity, thus hindering the advances of many applications. 

\begin{figure}[t]
\centering
\includegraphics[width=8.2cm]{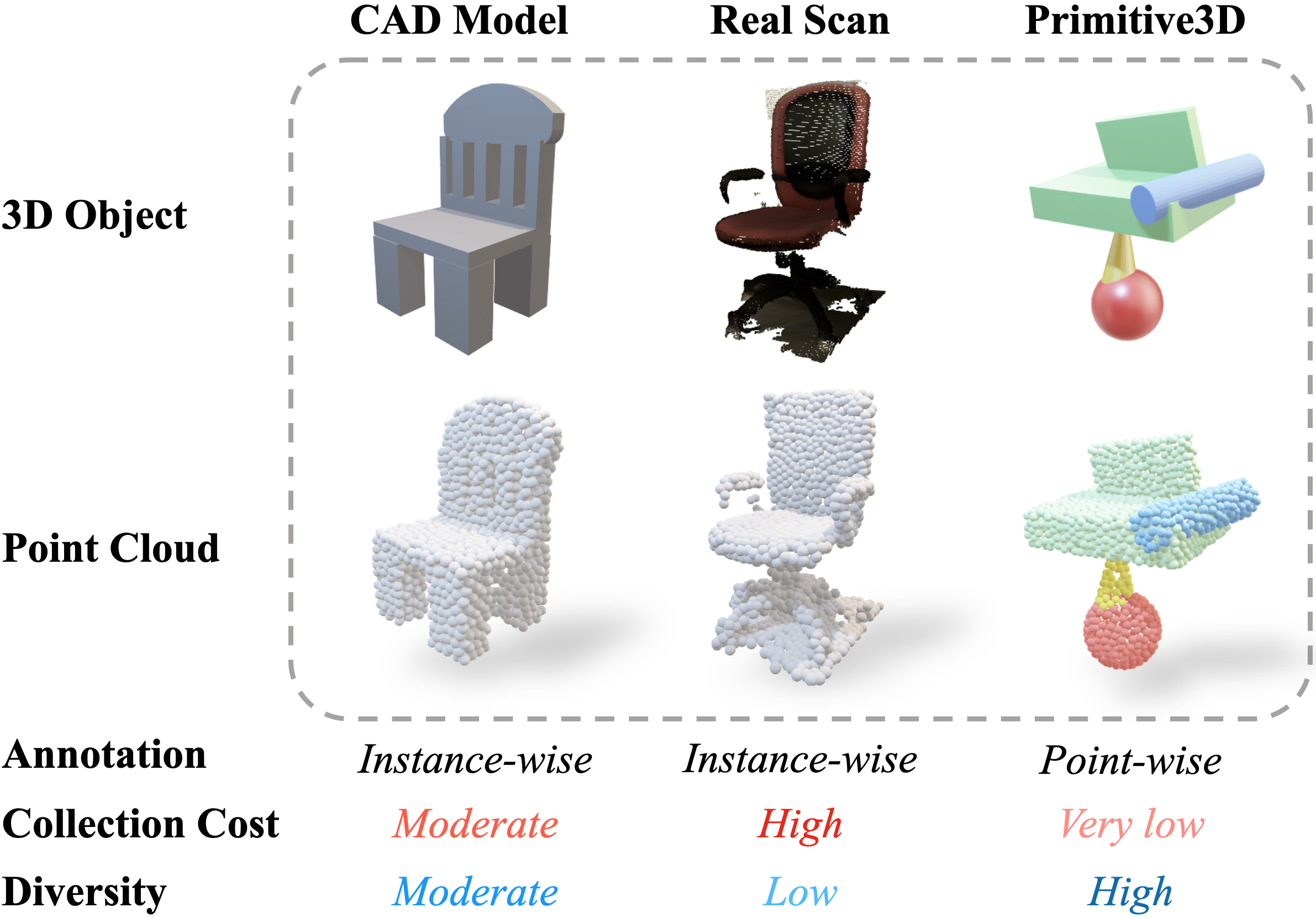}
\vspace{-0.1em}
\caption{Compared to the CAD model and real-world data, the Primitive3D dataset has richer annotations, lower collection cost, and higher sample diversity. These advantages help facilitate the learning of general knowledge for understanding 3D objects.}
\label{fig:compare}
\end{figure}

The main reason for the limitation of 3D object datasets is the substantial cost related to data collection and annotation.  In the current practice of real-world 3D object data acquisition, LiDAR or RGB-D scans often necessitate considerable labour and device costs~\cite{de2013unsupervised,dai2017scannet,hua2016scenenn}, while image-derived data requires much computation efforts~\cite{singh2014bigbird,li2020campus3d}. On the other hand, although synthesis datasets often consist of 3D CAD models with rich online data sources~\cite{shapenet2015,wu20153d, koch2019abc}, they still involve manual design works and also have limited generalization for real-world objects recognition\cite{uy2019revisiting}. 
Ultimately, tremendous human efforts have to be made in  the annotation and maintenance of large-scale labeled  datasets (\eg, voxel or point based label), especially considering the bulky size and the extra dimension of 3D data compared to 2D.
To alleviate these problems, attempts have been made to automatically generate 3D objects via generative models built from existing datasets~\cite{wu2016learning, khan2019unsupervised, nash2017shape, achlioptas2018learning}. Although these approaches are capable of generating high-quality 3D data with fewer human interactions, they are task-oriented, computationally intensive, and may potentially incorporate bias from the training dataset.
Therefore, it is critical to find a cheap 3D object data source as an alternative, which can derive large-scale, diverse and richly-annotated datasets, to push forward the development of 3D deep learning.

In our work, to obtain annotated 3D data at low cost, we introduce a learning-free method for the synthesis of pseudo 3D objects. Motivated by Constructive Solid Geometry (CSG) scheme~\cite{requicha1977constructive}, we build a vast number of random objects from basic 3D shapes, \ie, primitives. In a typical CSG scheme, 3D solids are often constructed with a tree representation, where the leaves are primitives and internal nodes are boolean set operations. We randomize this tree-based construction by setting the parameters including tree structure, primitive parameters, boolean operations and rigid transformations as random variables. By uniformly sampling such variables, we derive a random tree, and a random object can be obtained by executing this tree from bottom to top. By tracking the original primitives after the construction, the object is automatically annotated with part-based labels.
Furthermore, our analysis shows that such a method can generate objects with sufficient diversity, which allows the deep models to learn generalized representations from the resulting large-scale 3D dataset.

To exploit the abundance and annotations of the generated objects, we provide a multi-task learning method for the point cloud of such objects. To grasp the object geometry on a local and global level, the method combines two learning tasks: \emph{supervised segmentation} of point clouds and \emph{unsupervised reconstruction} of the original points.
Regarding the tremendous size of the generated dataset, we then present a method, dataset distillation,  that is integrated into the learning process to relieve the computational burden. This method is accomplished by eliminating certain samples from the generated dataset to shrink its maximum mean discrepancy (MMD) to a target dataset.

In experiments, we evaluate the validity of our method by multiple object classification benchmarks. The results of the cross-dataset classification show that the features learned from our dataset can surpass those learned from other widely used 3D object datasets.
Furthermore, pretraining by our method combined with fine-tuning can consistently improve the performance of downstream tasks with varying data size. Additionally, pretraining with dataset distillation can obtain comparable or even better performance than that with full dataset, reducing 86\% of the pretraining time.
In summary, our contributions are:
\begin{enumerate}
\vspace{-0.16cm}
\item We present a cost-efficient method to generate a large amount of valid and diverse random 3D objects with part annotations automatically. The generated dataset as well as generation scripts will be released publicly.
\vspace{-0.16cm}
\item We provide a multi-task learning method to train feature encoders on our generated dataset with dense annotations. We also propose an approach, dataset distillation,  that can be optionally employed in the learning process to lower the computation cost.
\vspace{-0.16cm}
\item 
Experiments show that our dataset serves as the best pretraining data for multiple downstream classification tasks in comparison to other commonly used datasets. Our pretraining method can also consistently help the downstream tasks in achieving higher performance. 
\vspace{-0.16cm}
\end{enumerate}

\section{Related Work}

\subsection{Dataset for 3D Object Understanding} 
In general, 3D object datasets can be divided into two categories: (1) \emph{synthesis dataset:} the  datasets of synthetic objects are often derived from CAD models, which can be manually designed or collected from websites. The most commonly used and general datasets are  ModelNet\cite{wu20153d} and ShapeNet\cite{shapenet2015}, as well as others for the particular usage, \eg, \cite{kim2020large, koch2019abc} for mechanical design. Despite the relatively large scale of these synthesis datasets, research indicates that models built from them may not necessarily generalize to recognize real objects\cite{uy2019revisiting}. (2) \emph{real-world 
dataset:} there are a few object datasets from real-world scans \cite{de2013unsupervised, uy2019revisiting}, but most are miniature in scale due to the high cost of the scanning and reconstruction process.
Besides the synthesis and real-world datasets, many works generate 3D shapes by deep models, like Deep Belief Networks\cite{wu20153d}, GAN\cite{achlioptas2018learning,wu2016learning}, VAE\cite{nash2017shape, wu2019sagnet} and Flow\cite{yang2019pointflow}. However, the generative models often suffer from considerable learning costs and the limitation of sample diversity.
In addition, most of these datasets are not point-wisely labeled, except for\cite{mo2019partnet, yi2016scalable}. Overall, point-wise annotations are difficult and costly to acquire, and the demand for high-quality 3D object datasets with extensive annotations and diversified data is still great.

\subsection{Deep Learning for 3D Object Understanding} 

\textbf{Deep Models.} Countless attempts have been made to apply deep learning models on a variety of 3D modalities, including voxels~\cite{maturana2015voxnet}, mesh~\cite{hanocka2019meshcnn}, multi-view~\cite{su2015multi}, and point cloud~\cite{qi2017pointnet,qi2017pointnet++}. Among these works, the point cloud-based method has received the most attention for its simplicity. Recently an increasing number of such models ~\cite{liu2019relation, boulch2020convpoint, wu2019pointconv, li2018pointcnn, wang2019dynamic} have shown their powers for 3D object understanding.

\textbf{Model Pretraining.} 
Different from model pretraining of 2D vision tasks~\cite{ding2020phraseclick,ding2021interaction,ding2021vision,ding2022deep,shuai2018toward,liu2019feature,liu2022instance,liu2021towards,mei2019deepdeblur}, recent researches on deep model pretraining for understanding 3D objects mainly focus on unsupervised learning methods. Typical unsupervised tasks include self-reconstruction~\cite{hassani2019unsupervised,yang2018foldingnet,gadelha2018multiresolution,han2019multi, achlioptas2018learning} and self-supervised learning pretext tasks~\cite{NEURIPS2019_993edc98, gao2020graphter,xie2020pointcontrast, wang2021unsupervised, rao2020global}. Most of these works rely on delicately designed  pretraining schemes rather than unlocking the potential of the pretraining dataset. In contrast, we pay more attention  to the automatic generation and annotation of pretraining datasets. 

\textbf{Domain Adaption.} Considering the domain gaps between different 3D object datasets, our work is also related to the field of domain adaption (DA) \cite{ben2007analysis, long2015learning, dai2021idm,dai2021dual}. Indeed, there are efforts to implement domain adaption in 3D objects understanding, such as PointDAN \cite{qin2019pointdan} and SSL-DA \cite{achituve2021self}. However, their approaches are proposed for model-centric DA, while we design our DA method from the perspective of the dataset, as specified in Section \ref{subsection:4.2}. 

\subsection{Primitives in 3D Deep Learning} 

Some latest researches have investigated decomposing or fitting 3D objects into primitive representations by deep learning approach~\cite{sharma2018csgnet,li2019supervised,tulsiani2017learning, deng2020cvxnet,paschalidou2019superquadrics, yavartanoo20213dias}. Other methods related to primitives focus on deep shape generation models, where the primitives act as intermediate representations or building blocks~\cite{wu2020pq,zou20173d,khan2019unsupervised}. These works demonstrate the representation ability of assembled primitives, which inspires our usage of primitives to generate 3D object data.

\section{Dataset Construction}

\subsection{Data Generation with Random Primitives}
\label{subsection:3.1}
We start with the building blocks of the CSG scheme, namely, the primitives. In CSG-based modeling,  the primitive is defined by a type and a set of parameters. The primitive type $\Psi_i$ 
is a basic shape, such as a box, sphere, cylinder, cone, and etc~\cite{requicha1977constructive}. Each type $\Psi_i$ contains a collection of objects $\psi_{\theta}$ with the same shape in the canonical form, say centered at an original point and bounded by a unit ball. Specifically, a canonical instance $\psi_{\theta}$ is parameterized by $\theta \in \Theta^{\Psi_i}$, for example, the height, width, and depth of a box. 
Moreover, a particular primitive instance $\psi'_{\theta}$ can be obtained from $\psi_{\theta}$ by\vspace{-0.2cm}\begin{equation}
\label{eq:transform}
\psi'_{\theta}= \lambda(\Phi\psi_{\theta}) + \delta,
\vspace{-0.2cm}\end{equation} where $\Phi\in SO(3)$ is a rotation transformation, $\delta\in\mathbb{R}^3$ is a translation vector and $\lambda \in \mathbb{R}^{+}$ is a scaling factor.

Based on the primitive instances, we design a randomized method to create complicated pseudo objects via the CSG scheme.
CSG construction is often represented as a binary tree in which the leaf nodes are primitive instances, while the internal nodes are boolean set operations applied to the immediate children\cite{requicha1977constructive}.
To obtain random objects, our method is to randomize such a construction process as \textit{Randomized Constructive Tree (RCT)}. The RCT treats each parameter, including the tree structure, as a random variable, such as a uniform distribution, see Algorithm~\ref{alg:gen}.
Particularly, we first specify the domain of each RCT parameter, including a finite set of primitive type $\mathbb{P}=\{\Psi_1,\cdots, \Psi_p\}$, a collection of parameter sets  $\{\Theta^{\Psi_1},\cdots, \Theta^{\Psi_p},\Lambda\}$ where each element is compact, and a set of boolean operations $\mathbb{O}$. Such setups allow us to sample each parameter uniformly.
To sample the tree structure with $l$ leaves uniformly, we obtain the binary tree $\tau_{2l-1}$ by R\'{e}my's algorithm~\cite{alonso1997linear}. In addition, a sampled translation $\delta$ is applied to the left child of the internal node, ensuring that the boolean operation can hardly return an \textit{empty} object.

Compared to learning-based data generation approaches, we simply sample each parameter uniformly within the feasible set, avoiding learning costs and bias from the learned dataset.
In this randomized manner, we could generate an unlimited number of distinct pseudo 3D objects at low cost. We refer to the set of these objects as \textit{Primitive3D}. Moreover, because it is easy to track each part of the object to the original primitive, part-based annotations can be created automatically based on the primitive types and instances.

\vspace{-1em}
\begin{algorithm}
\footnotesize

\caption{Data Generation}
\begin{algorithmic}[1]\label{alg:gen}
\renewcommand{\algorithmicrequire}{\textbf{Input:}}
\renewcommand{\algorithmicensure}{\textbf{Output:}}
\REQUIRE leaf number $l$, type set $\mathbb{P}$, parameters set $\{\Theta\}$, scale range $\Lambda$\\
\quad \Hquad boolean operation set $\mathbb{O}$
\ENSURE Object $\tilde{\psi}_{l}$
\STATE $\tau_{2l-1} \leftarrow$ \textit{RandomBinaryTree}($2n-1$)
\FOR {$E_i$ \textbf{in} leaves of $\tau_{2n-1}$}

\STATE sample primitive: $\Psi_i \sim $ Uniform$(\mathbb{P})$, $\theta_i \sim  $ Uniform$(\Theta^{\Psi_i})$

\STATE sample transform: $\Phi_i\sim $ Uniform$(SO(3))$, $\lambda_i\sim $ Uniform$(\Lambda)$
\STATE  $\psi_{\theta_i} \leftarrow$ generate $(\Psi_i, \theta_i)$
\STATE $E_i \leftarrow \lambda_i(\Phi_i\psi_{\theta_i})$ 
\ENDFOR
\FOR {$I_j$ \textbf{in} internal nodes of $\tau_{2n-1}$ by \textit{bottom-up}}
\STATE sample operation: $\circ \sim $ Uniform$(\mathbb{O})$
\STATE sample point from child: $p_j^l\sim  $ Uniform$(\psi_j^l)$, $p_j^r \sim  $ Uniform$(\psi_j^r)$ 

\STATE $\delta_j\leftarrow {p_j}^r - {p_j}^l$
\STATE $I_j \leftarrow$ execute $ (\psi_j^r) \circ (\psi_j^l + \delta_j)$

\ENDFOR
\STATE $\tilde{\psi}_{l} \leftarrow$ root node of $\tau_{2l-1}$
\end{algorithmic}

\end{algorithm}

\vspace{-1.6em}

\subsection{Analysis of Randomized Constructive Tree}
\label{subsection:3.2}

In this subsection, we illustrate the capability of RCT by its closure property and approximability to arbitrary 3D objects. 3D objects in solid modeling are often defined as bounded, closed, regular, and semi-analytic subsets of $\mathbb{R}^3$, and their sets are referred to as \textit{r-sets}\cite{requicha1977mathematical}. Furthermore, the \textit{regularized boolean operations} are also adapted from the conventional ones for processing r-sets\cite{requicha1977mathematical}. We note the following property of RCT with above definitions.

\vspace{-0.5em}
\begin{mylem}
[Closure Property of RCT]\label{generality} If the leaf nodes of an RCT are r-sets and the internal nodes are regularized boolean operations, the sample of such RCT always corresponds to an r-set.

\end{mylem}
\vspace{-0.5em}

The Lemma~\ref{generality} ensures that every RCT sample can produce a valid 3D object when using r-set primitives, which is because the class of  r-set is closed under regularized operations~\cite{hironaka1975triangulations}. Additionally, RCT samples associated with properly defined primitive types can approximate any complex object. 
Such property suggests the sufficient diversity of RCT samples when conducting random sampling. We note that Appendix A.2 has a detailed description of this approximability as well as more definitions related to the r-sets theory.
The above properties of RCT enable us to generate massive Primitive3D objects in a random fashion while meeting both the volume and diversity demands.

\begin{figure*}[ht]
\vspace{-1.5em}
\centering
\includegraphics[width=17.5cm]{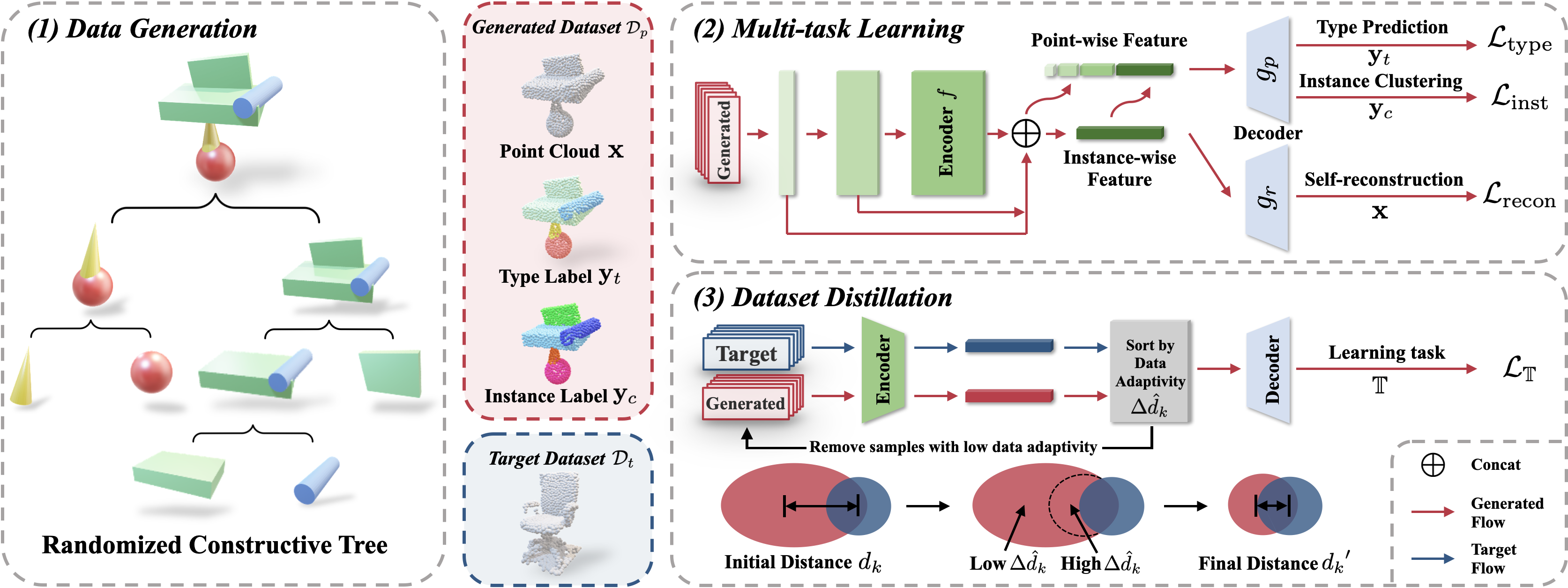}
\vspace{-1.2em}
\caption{Data generation of Primitive3D and its learning.}
\vspace{-0.5em}
\label{fig:framework}
\end{figure*}

\section{Learning from Primitive3D}
\label{subsection:4.1}

To take advantage of Primitive3D's annotation and enormous volume, we provide a multi-task learning method on the point clouds of Primitive3D objects, as shown in Figure~\ref{fig:framework}. Also, a dataset distillation method is designed for incorporation into the learning.

\subsection{Multi-task Learning}
A point cloud dataset $\mathcal{D}_p$ can be constructed by collecting $N$ sampled points $\mathbf{x}\in\mathbb{X}=\mathbb{R}^{N\times3}$ from  the surface of each Primitive3D object.
Each point in the point cloud is associated with a semantic label $\mathbf{y}_t$ and an instance label $\mathbf{y}_c$,  which describe the primitive type and instance information, respectively.
On the basis of such labels, we propose a multi-task learning that combines multiple levels of information to train a shared encoder $f$. The learning procedure is depicted in Figure~\ref{fig:framework}, while the  associated tasks are summarized in Table~\ref{tab:pretrain_tasks}. Then we detail each learning task.

\begin{table}[h]
\centering
\begin{footnotesize}

\vspace{-0.8em}

\begin{tabular}{lccc}

\toprule
Training task &Data description               & Groundtruth    \\ \hline
Semantic segmentation  &Primitive type label    &  $\mathbf{y}_t$                      \\
Instance segmentation &Primitive instance label&    $\mathbf{y}_c$                     \\
Reconstruction          &Point cloud      &   $\mathbf{x}$                       \\ \bottomrule
\end{tabular}
\vspace{-1em}
\caption{Learning tasks on Primitive3D.}
\label{tab:pretrain_tasks}
\vspace{-0.5em}
\end{footnotesize}
\end{table}

We define two supervised segmentation tasks based on the point-wise labels $\mathbf{y}_t$ and $\mathbf{y}_c$. Specifically, a point-based decoder $g_p$ is first employed on the point-wise feature output by the encoder. The segmentation tasks are conducted on the embeddings of $g_p$, with the loss being the summation of the losses of two branching tasks as:
\begin{equation}
\small
\label{eq:construct_loss}
\mathcal{L}_{\text{seg}} = \mathcal{L}_{\text{type}} + \alpha \cdot \mathcal{L}_{\text{inst}},
\end{equation}where $\alpha$ is a tunable weight, $\mathcal{L}_{\text{type}}$  is the cross-entropy loss for semantic segmentation of predicting $\mathbf{y}_t$, and $\mathcal{L}_{\text{inst}}$  is the discriminative loss for instance segmentation used by \cite{ pham2019jsis3d, wang2019associatively}. Instead of predicting the instance masks, $\mathcal{L}_{\text{inst}}$ aims to learn point-based embeddings for the class-agnostic clustering, where the ground truth of each cluster is a primitive instance. 

Self-reconstruction is implemented as the unsupervised task based on autoencoder structure, with the goal of reconstructing the original points. Particularly, a folding-based decoder\cite{yang2018foldingnet} $g_r$ is adapted to deform the canonical 2D grid into new 3D point coordinates $\mathbf{x}'$ conditioned on the global feature output from the encoder $f$. The task loss is defined as the augmented Chamfer distance between the input $\mathbf{x}$ and the reconstructed $\mathbf{x}'$ in  Equation~\eqref{eq:augmentedcd}, which is more robust under some ill cases\cite{chen2019deep} than the original version.
\begin{small}
\begin{equation}\label{eq:augmentedcd}
\mathcal{L}_{\text{recon}} =\max \left(\sum _{x_{i} \in \mathbf{x}}\min_{x_{j} '\in \mathbf{x}'}\Vert x_{i} -x_{j} '\Vert _{2}, \sum _{x_{j} '\in \mathbf{x}'}\min_{x_{i} \in \mathbf{x}}\Vert x_{i} -x_{j} '\Vert _{2}\right).
\end{equation}
\end{small}
Overall, the total loss is the weighted sum of all task losses \begin{equation}
\small
\mathcal{L} =\mathcal{L}_{\text{seg}} + \beta \cdot \mathcal{L}_{\text{recon}}.
\end{equation}
We note that the segmentation task enables supervised training of the model's ability to comprehend parts of Primitive3D object, while the unsupervised learning guarantees that the model retains global geometric knowledge.
Their combination allows the learning of generalized feature representations from the Primitive3D dataset that can be used for both local and global understanding of 3D objects. 

\subsection{Dataset Distillation}
\label{subsection:4.2}

Although learning on a large-scale dataset conceivably produces more general feature representations, the computational cost of such a learning process is also considerable. To improve the learning efficiency, we propose a method for reducing the Primitive3D data in order to learn only the features associated with a target dataset. This process of reducing the learning data according to the target dataset is referred to as \textit{dataset distillation}.

\textbf{A Bound from Domain Adaption.} 
To motivate our method of dataset distillation, we first introduce a learning bound from the domain adaption theory.
Let $\mathcal{D}_p$ and $\mathcal{D}_t$ denote the generated dataset for the feature learning task and the target dataset, in which the data are sampled $i.i.d$ from domain $\mathcal{P}$ and $\mathcal{T}$, respectively.
When the data domains $\mathcal{P}$ and $\mathcal{T}$ are distinct but relevant, according to the theory in \cite{ben2007analysis,long2015learning}, by defining a hypothesis $h\in \mathcal{H}$,  the expected risk of $h$ on target domain $\epsilon_{\mathcal{T}}(h)$ can be bounded by the expected risk $\epsilon_{\mathcal{P}}(h) $ on the $\mathcal{P}$ and the distance between $\mathcal{P}$ and $\mathcal{T}$, \ie, \vspace{-0.06cm}\begin{equation}
\small
\label{eq:upper_bound}
\epsilon_{\mathcal{T}}(h) \leqslant \epsilon_{\mathcal{P}}(h) + 2d_k(\mathcal{P}, \mathcal{T}) + C_0,
\vspace{-0.1cm}\end{equation}where $C_0$ is a constant associated with $\mathcal{H}$, and $d_k$ is the maximum mean discrepancy (MMD) between two domains.
As is well-known in \cite{vapnik1998statistical}, decreasing the scale of the dataset $\mathcal{D}_p$ will increase the empirical upper bound of expected risk $\epsilon_{\mathcal{P}}(h)$ generally, thus raise the bound of $\epsilon_{\mathcal{T}}(h)$.
To compensate such growth,  the second term $d_k$ in Equation~\eqref{eq:upper_bound} should be 
suppressed when removing samples from $\mathcal{D}_p$, which motivates our strategy of dataset distillation.
In the following, we define a new metric, \textit{data adaptivity}, to remove samples from $\mathcal{D}_p$ in order to reduce the  distance $d_k$. Then we propose our dataset distillation based on it.

\vspace{0.3em}
\textbf{Distillation by Data Adaptivity.} Our definition of data adaptivity is simply the difference between the dataset distances before and after removing a subset of samples
\begin{myDef}
[Data Adaptivity] Let $\mathcal{D}$ and $\mathcal{D}'$ denote two datasets, and $d(\cdot, \cdot)$ is a metric that measures the distance between two datasets. The $d$-based data adaptivity of a subset $X\subseteq \mathcal{D}$ with respect to $\mathcal{D}'$ is defined as
\begin{small}
\begin{equation}
\label{eq:data_adaptivity}
\Delta_{d}(X;\mathcal{D},\mathcal{D}') \triangleq d\left( \mathcal{D}\backslash X,\mathcal{D}'\right) -d( \mathcal{D},\mathcal{D}').
\end{equation}
\end{small}
\end{myDef}
This definition quantifies the impact of subtracting $X$ from $\mathcal{D}$ on the value of $d(\mathcal{D}, \mathcal{D}')$. Therefore, we define the dataset distillation problem that is to find a subset $X$ that \begin{equation}
\small
    \max_{X\subseteq \mathcal{D}} \Delta _{d}(X;\mathcal{D},\mathcal{D}') \ \text{s.t.}\  |X|=n',
\end{equation}
where $n'$ is the desired size of the distilled dataset. In general, solving this problem is very difficult because of the large searching space. In practice, we consider an approximated approach such that the importance of each sample $\mathbf{x}\in\mathcal{D}$ is measured by $\Delta_{d}(\mathbf{x};\mathcal{D}, \mathcal{D}')$\footnote{With an abuse of notation, $\Delta_{d}(\mathbf{x};\mathcal{D}, \mathcal{D}')$ is indeed $\Delta_{d}(\{\mathbf{x}\};\mathcal{D}, \mathcal{D}')$} and we remove the samples with less relative importance.

Now we introduce an efficient method to sort data adaptivity in practice when $d$ is specified as MMD. Suppose $\mathbf{x}^p\in\mathcal{D}_{p}$ and $\mathbf{x}^t\in\mathcal{D}_{t}$, according to \cite{gretton2012kernel}, an empirical estimation  of $d_k^2(\mathcal{P}, \mathcal{T})$  is given by
\begin{equation}
\footnotesize
\begin{array}{ll}
\hat{d}_{k}^{2}(\mathcal{D}_{p},\mathcal{D}_{t})\triangleq&\frac{1}{m^{2}}\sum\limits_{i,j} k\left(\mathbf{x}_{i}^{p} ,\mathbf{x}_{j}^{p}\right) -\frac{2}{mn}\sum\limits_{i,j} k\left(\mathbf{x}_{i}^{p} ,\mathbf{x}_{j}^{t}\right) 
\\&+\frac{1}{n^{2}}\sum\limits_{i,j} k\left(\mathbf{x}_{i}^{t} ,\mathbf{x}_{j}^{t}\right)
\end{array}
\end{equation} 
where $k$ is the combination of multiple Gaussian Radial Basis Function (RBF) kernels, while $m$ and $n$ are the size of $\mathcal{D}_{p}$ and $\mathcal{D}_{t}$. By substituting  it to Equation~\eqref{eq:data_adaptivity}, the MMD-based data adaptivity  $\Delta_{\hat{d}_k}(\mathbf{x}^p;\mathcal{D}_p,\mathcal{D}_t)$ is properly defined and can be explicitly computed for given $\mathcal{D}_p$ and $\mathcal{D}_t$. However, naively computing and sorting $\Delta_{\hat{d}_k}(\mathbf{x}^p;\mathcal{D}_p,\mathcal{D}_t)$ for all $\mathbf{x}^p$ is costly, thus we propose in Lemma ~\ref{theo:approx} a proxy that can be calculated efficiently and achieve the same sorting result for $\Delta_{\hat{d}_k}$ under a mild condition.
\vspace{-0.03in}
\begin{mylem}
\label{theo:approx}
Let $\mathcal{D}=\{\mathbf{x}_i\}_{i=1}^{m}$ and $\mathcal{D}'=\{\mathbf{x}'_j\}_{j=1}^{n}$ denote two distinct datasets, and $\Delta_{\hat{d}_k}(\mathbf{x};\mathcal{D},\mathcal{D}')$ denotes the MMD-based data adaptivity of one data $\mathbf{x}\in\mathcal{D}$. With the assumption that $	\left|\Delta_{\hat{d}_k}(\mathbf{x};\mathcal{D},\mathcal{D}')	\right| \ll \hat{d}_k(\mathcal{D}, \mathcal{D}')$, sorting the data adaptivity $\Delta_{\hat{d}_k}(\mathbf{x};\mathcal{D}, \mathcal{D}')$ can be achieved by sorting the following proxy quantity\vspace{-0.06in}
\begin{equation}
\small
\label{eq:est_data_apapt}
\frac{1}{n}\sum\limits_{\mathbf{x}_j\in\mathcal{D}'} k\left( \mathbf{x} ,\mathbf{x}_{j}'\right) -\frac{1}{m-1}\sum\limits_{\mathbf{x}_i\in\mathcal{D}} k\left( \mathbf{x} ,\mathbf{x}_{i}\right).
\vspace{-0.03in}
\end{equation}
\end{mylem}

The idea of this lemma is intuitive: data adaptivity of $\mathbf{x}^p$ depends on how close it is to $\mathcal{D}_t$ and how far it is from $\mathcal{D}_p$ indicated by $k$. By calculating such proxy, the computation complexity for sorting  data adaptivity is reduced dramatically from $O(2m(m+n)^2)$ to $O(m(m+n))$.

\vspace{-1em}
\begin{algorithm}[h]
\footnotesize
\caption{Data Distillation}
\begin{algorithmic}[1]
\label{alg:distillation}
\renewcommand{\algorithmicrequire}{\textbf{Input:}}
\renewcommand{\algorithmicensure}{\textbf{Output:}}
\REQUIRE  model $f$, learning task $\mathbb{T}$, learning task loss $\mathcal{L}_{\mathbb{T}}$, ratio $r$\\ threshold $size_{t}$,
  epoch $L$, generated dataset $\mathcal{D}_p$, target dataset $\mathcal{D}_t$ \\

\ENSURE trained model $f_L$
\STATE $size\leftarrow |\mathcal{D}_p |$, Initialize $f_0$
\FOR {$i = 1,\cdots, L$}
\STATE $f_i \leftarrow$ train $f_{i-1}$ on $\mathcal{D}_p$ by task $\mathbb{T}$ with loss $\mathcal{L}_{\mathbb{T}}$
\IF{$size > size_{t}$}
\STATE $size \leftarrow \max(\lfloor r\cdot size\rfloor, size_{t})$

\STATE Sort $\mathcal{D}_p$ by $\Delta_{\hat{d}_k}(\mathbf{x}^p;\mathcal{D}_p,\mathcal{D}_t)$ in descending order

\STATE $\mathcal{D}_p \leftarrow \mathcal{D}_p [1,\cdots, size]$
\ENDIF
\ENDFOR
\RETURN $f_L$
\end{algorithmic}
\end{algorithm}
\vspace{-1.2em}

Putting everything together, we propose a workflow of progressively pruning the data of lower data adaptivity from $\mathcal{D}_p$ as illustrated in Algorithm~\ref{alg:distillation}, where $\mathbb{T}$ is our learning task on Primitive3D,  $size_t$ is a threshold deciding the size of the distilled dataset and $r$ is the retention ratio in each distillation step. The parameter setup for $r$ and $size_t$ is specified in Section~\ref{subsection:5.3}.
Similar to previous work\cite{tan2017distant}, we calculate the kernel function using the global features learned from self-reconstruction rather than the original $\mathbf{x}$.
With the proposed method, the learning data from Primitive3D can be substantially less than the original, markedly cutting off the computational overhead without losing the efficacy of features for the target dataset as experiments show.

~~~~

\begin{table*}[t]
\centering

\begin{tabular}{l|ccc|ccc|ccc}
\toprule
\multirow{2}{*}{Pretraining dataset}   
     & \multicolumn{3}{c|}{ModelNet40} & \multicolumn{3}{c|}{ScanObjectNN} & \multicolumn{3}{c}{ScanNet10} \\
    & \ ~~Uns~~      &  ~~Sup~~         & Uns~\&~Sup      & ~~Uns~~       & ~~Sup~~          & Uns\&Sup      & ~~Uns~~      & ~~Sup~~         & Uns\&Sup     \\\hline
ModelNet40         &-        & -         & -         &68.9         & 70.1       & 72.5      &71.9        & 64.4      & 70.0    \\
ScanObjectNN       &85.3        & 85.2      &84.6       &-         & -          & -         &67.4        & 62.0        &68.0    \\
ScanNet10          &86.3        & 85.3      & 85.6      &70.7         & 68.5       & 70.2      &-        & -         & -        \\
ShapeNet           & \underline{87.1}       & 87.3      & 88.0      &\underline{72.2}         & 73.2       &74.5       &\underline{72.3}        & 66.3      & 69.3     \\
Primitive3D (Ours) &86.1        & \underline{88.9}      & \underline{\textbf{89.4}}      &70.8         & \underline{76.7}       & \underline{\textbf{78.5}}      &70.0         & \underline{72.0}      & \underline{\textbf{72.9}}  \\  \bottomrule
\end{tabular}
\vspace{-0.5em}
\caption{The comparison of cross-dataset classification accuracy (\%) on various 3D datasets. ``Sup'' and ``Uns'' denotes the pretraining with only supervised task and unsupervised task, respectively, while ``Uns\& Sup'' denotes the pretraining with the combination of them. 
}
\vspace{-0.2em}
\label{tab:svm_acc_compare}
\end{table*}

\section{Experiments}

\subsection{Experiments Setup}
\label{subsection:5.1}

\textbf{Dataset Preparation.}
We first generate a set of 150,000 point clouds extracted from  Primitive3D objects sampled by the RCT approach proposed in Section~\ref{subsection:3.1}. The point cloud dataset is named Primitive3D in this section, which is generated once and fixed throughout the experiments except for the ablation study on dataset size in Section~\ref{subsection:5.3}. To achieve a good trade-off between object complexity and generation time , the number of leaves in RCTs is chosen in the range of 1 to 6  for our generation as detailed in Section~\ref{subsection:5.3}. In our implementation, the boolean operations set $\mathbb{O}$  contains only the regularized union, which is sufficient to generate complex objects and is easy to implement. The primitive types $\mathbb{P}$ of RCTs include five shapes, namely, sphere, box, cylinder, cone, and torus. 

\begin{table}[h]
\vspace{-0em}
\centering
\scalebox{0.82}{
\begin{tabular}{lcccc}
\toprule
Dataset      & Initialism & Type      & \#Class & Split     \\ \hline
ModelNet40\cite{wu20153d}    & MN40       & CAD & 40      & 9,840 / 2,468 \\
ScanObjectNN\cite{uy2019revisiting} & SONN       & Real  & 15      & 2,309 / 581  \\
ScanNet10\cite{qin2019pointdan}    & SN10       & Real  & 10      & 6,110 / 1,769 \\
ShapeNet\cite{shapenet2015}     & SN         & CAD & 55      & 52,472 / -   \\ \bottomrule
\end{tabular}}
\vspace{-0.3cm}
\caption{3D object datasets for classification task.}
\label{tab:datasets}

\end{table}

\textbf{Network Architecture.} Unless otherwise specified, we apply the widely used DGCNN \cite{wang2019dynamic} as the encoder-decoder network in the experiment. The unsupervised reconstruction task employs a decoder adapted from \cite{yang2018foldingnet} which takes the 1024-dimensional encoded feature as input, while the two supervised segmentation tasks employ a shared point-wise decoder with the same architecture as the official implementation of point cloud semantic segmentation. The number of input points is 1024 except for tasks involving object part segmentation, which uses the number of 2048.

\textbf{Preraining Setting.} We employ the multi-task learning  described in Section~\ref{subsection:3.1} to pretrain deep encoders on Primitive3D to obtain feature representations. The training optimizer is set to Adam without weight decay. We train the model for 50 epochs, beginning with a learning rate of 0.001, and decaying it by 0.7 every 10 epochs. For all training processes in the following content, the batch size is set to 32.  In our multi-task learning, the parameters of discriminative loss for the instance segmentation are identical to the original configuration\cite{de2017semantic}, while the weights $\alpha=0.05$ and $\beta=0.2$, respectively. For dataset distillation, we set $r=0.7$ and $size_t=10000$ except for the ablation study.

\subsection{Main Results}
\label{subsection:5.2}

We first pretrain the models by the multi-task learning on Primitive3D. The output feature representations are then evaluated on multiple object classification benchmarks and compared to the models pretrained on other datasets.
We consider four commonly used 3D object datasets for our experiments, namely, ModelNet40~\cite{wu20153d}, ScanObjectNN~\cite{uy2019revisiting}, ScanNet10~\cite{qin2019pointdan} and ShapeNet\cite{shapenet2015}, with the statistics summarized in Table~\ref{tab:datasets}. The first three datasets are utilized for both benchmarking and pretraining, while the ShapeNet is included only to serve the pretraining purpose. It is worth mentioning that, compared to the synthesis dataset ModelNet40, the classifications on ScanObjectNN and ScanNet are more challenging due to the presence of real-world noise such as the occlusion of objects. In our comparison, we use the same unsupervised pretraining task on the compared datasets, while different supervised pretraining tasks based on their own annotations. Finally, identical settings are applied to all pretraining tasks.   

\textbf{Cross Dataset Evaluation.} 
We conduct cross data evaluation on the benchmark datasets. Specifically, we pretrain the feature encoders using supervised and/or unsupervised learning methods on the pretraining dataset. We test the efficacy of the output features by using a linear SVM trained on the benchmark datasets to perform classification.
The results in Table~\ref{tab:svm_acc_compare} show that the supervised pretraining itself on Primitive3D is enough to outperform other commonly used datasets, except that the performance on ScanNet10 is slightly behind. Moreover, adding in unsupervised pretraining guarantees the superiority of Primitive3D against all compared datasets. We also note that the unsupervised and supervised pretraining on Primitive3D always enhance each other, while on the other datasets their combination sometimes hurts the performance. This suggests that our part-based annotations might be more appropriate to guide the geometry learning of 3D objects.

\begin{table*}[h]
\centering
\vspace{-0.3em}
\scalebox{1}{
\begin{tabular}{l|c|cccc|cccc}
\toprule
\multicolumn{1}{l|}{\multirow{2}{*}{Model Pretraining}}      &Pretraining  & \multicolumn{4}{c|}{\# Training Samples of ModelNet40} & \multicolumn{4}{c}{\# Training Samples  of ScanNet10} \\
\multicolumn{1}{c|}{}   &time  & ~~~400~~~    & ~~~800~~~    & ~1200~~   & All  & ~~~400~~~    & ~~~800~~~    & ~1200~~  & All  \\ \hline
Random Init    & -  &73.2        &79.4        &84.3   & 92.0      &66.4        &69.9        &71.8    & 77.7      \\
Primtive3D (Random Drop) & 12\%        &76.2        &80.0        &83.2   & 91.4      &67.3        &67.8        &69.9     & 76.9   \\

Primtive3D (Distillation)  & 14\%    &\textbf{78.0}        &\textbf{82.1}        &84.7     & 92.0    &\textbf{68.9}        &70.9        &72.6     & 77.9    \\
Primtive3D (Full)  & 100\%    &77.1        &82.0       &\textbf{85.7}    & \textbf{92.1}     &68.4        &\textbf{71.8}        &\textbf{73.1}    & \textbf{78.1}     \\
\bottomrule
\end{tabular}}

\caption{Classification accuracy (\%) on ModelNet40 and ScanNet10 test sets with various training sample numbers. The comparisons are based on different model initialization:  \textbf{Random Init:} randomly initialized; \textbf{Random Drop:} pretrained on randomly dropped Primitive3D which is indeed Algorithm~\ref{alg:distillation} with random shuffling in step 6; \textbf{ Distillation}: pretrained with the proposed dataset distillation on Primitive3D; \textbf{Full}: pretrained on full Primitive3D. The pretraining times are reported as the relative ratio of full Primitive3D pretraining time. }
\label{tab:semi_learning}
\end{table*}

\textbf{Fine-tuning with Varying Data Size.} 
To check whether our method could boost the downstream task performance, we use the pretrained weights by our learning method to initialize the deep classifier and fine-tune it on benchmark datasets. Besides using the full benchmark datasets, the fine-tuning is also conducted on partial training data to evaluate the data efficiency of models.
We imply the same training settings as the official implementation\cite{wang2019dynamic} for all runs, except that a lower initial learning rate of 0.01 is applied during fine-tuning procedure.  
It can be observed from Table~\ref{tab:semi_learning} that our method surpasses the train-from-scratch method for either synthesis or real-world datasets, especially when the training samples are limited.
Furthermore, we report the result of dataset distillation.  For the dataset distillation method, the time reported is for the ModelNet40 dataset of 1200 samples, which is the maximum among all six pretraining times. Compared to the full Primitive3D pretraining, our distillation method saves 86\% of the time, with at most 1\% performance degradation for all tasks. Additionally, dataset distillation substantially outperform the train-from-scratch and the random drop method, especially when the data scale is less than 800 it also exceeds full Primitive3D pretraining.

\subsection{More Studies}
\label{subsection:5.3}
\begin{figure}[hbt!]
    \centering
   \begin{subfigure}[Primitive3D]{2.72cm}
        \centering
       \includegraphics[width=2.72cm]{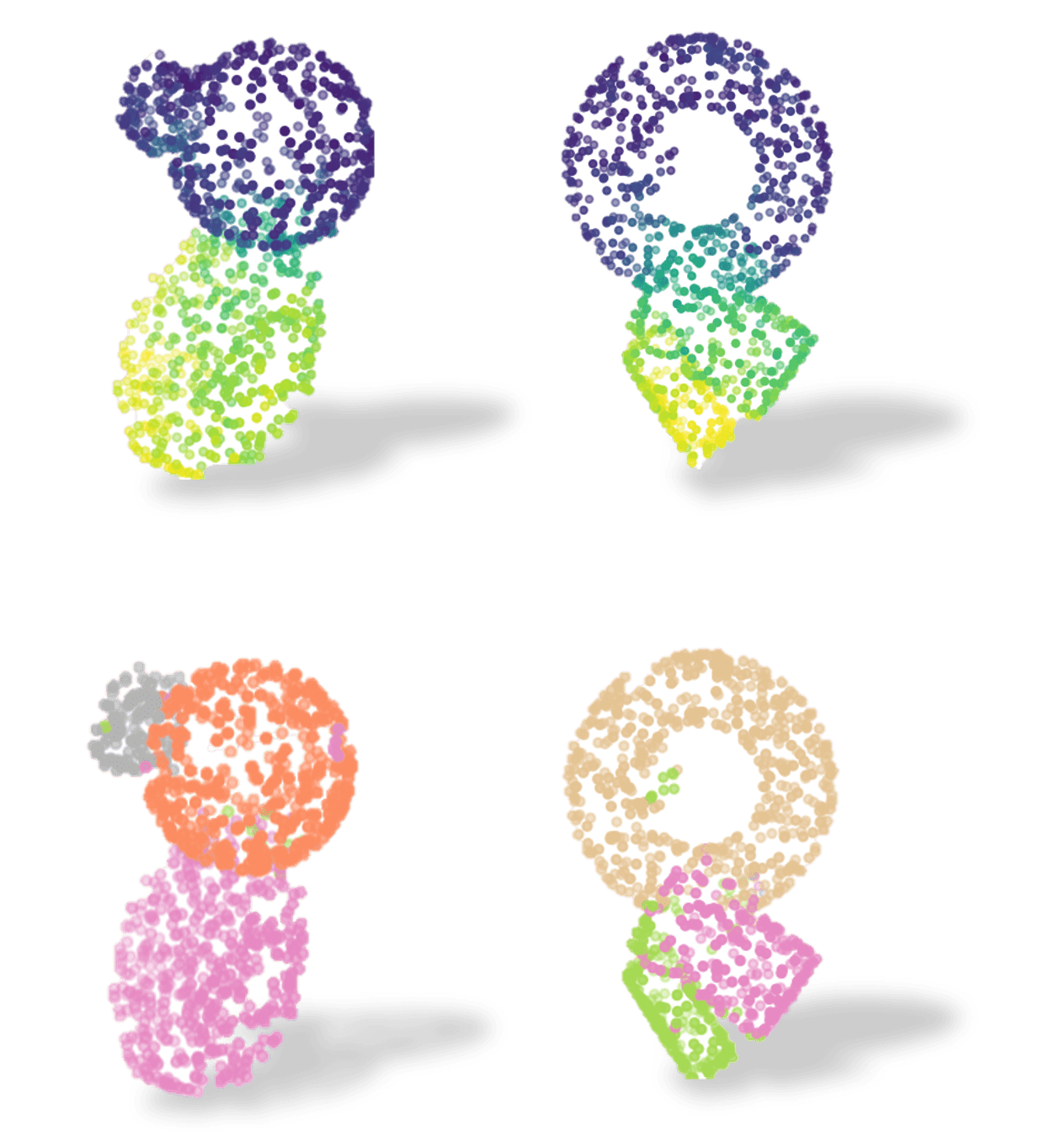}

        \caption{Primitive3D}

    \end{subfigure}
   \begin{subfigure}[ModelNet40]{2.72cm}
        \centering
       \includegraphics[width=2.72cm]{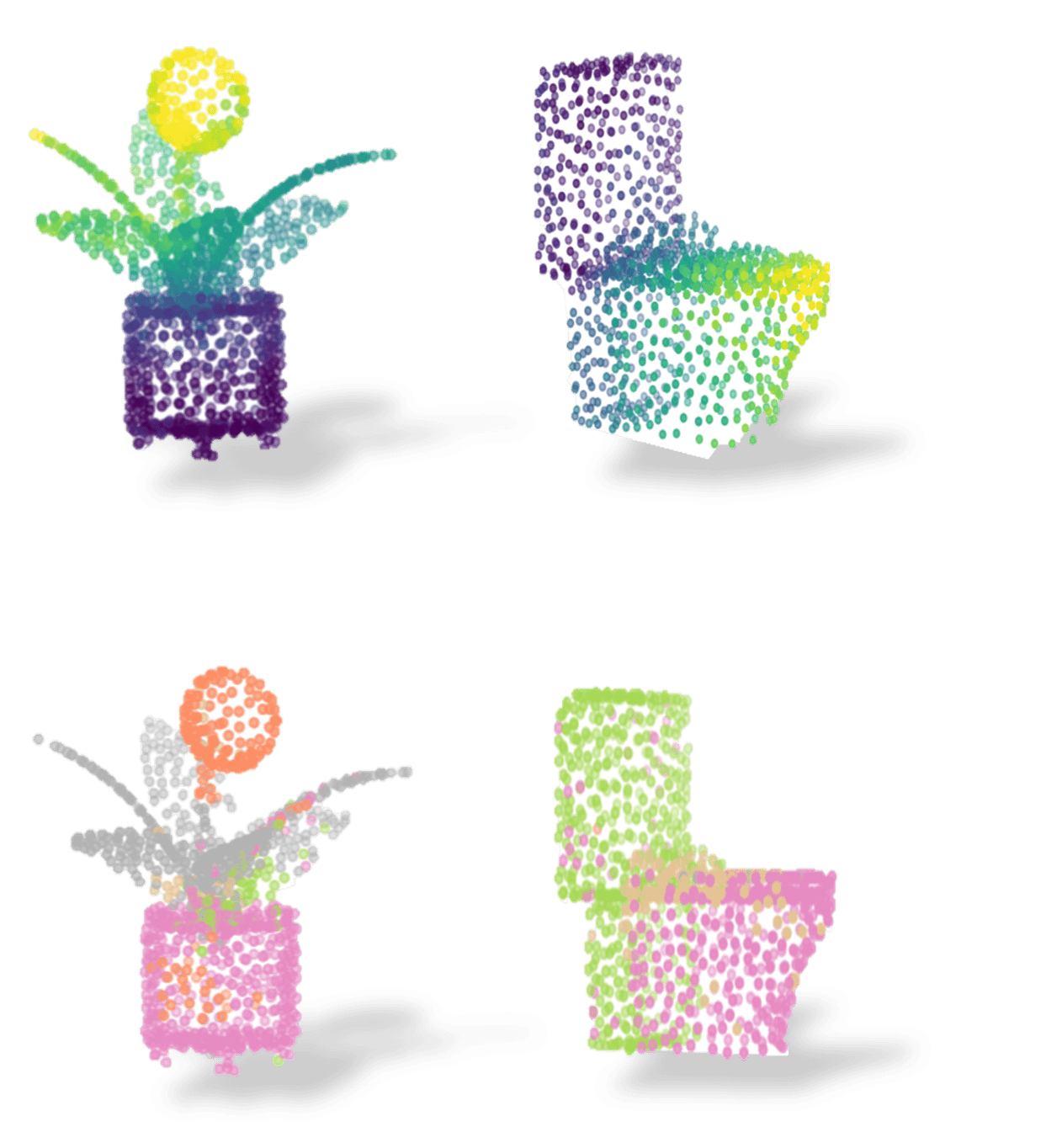}
        \caption{ModelNet40}
    \end{subfigure}
       \begin{subfigure}[ScanNet10]{2.72cm}
        \centering
       \includegraphics[width=2.72cm]{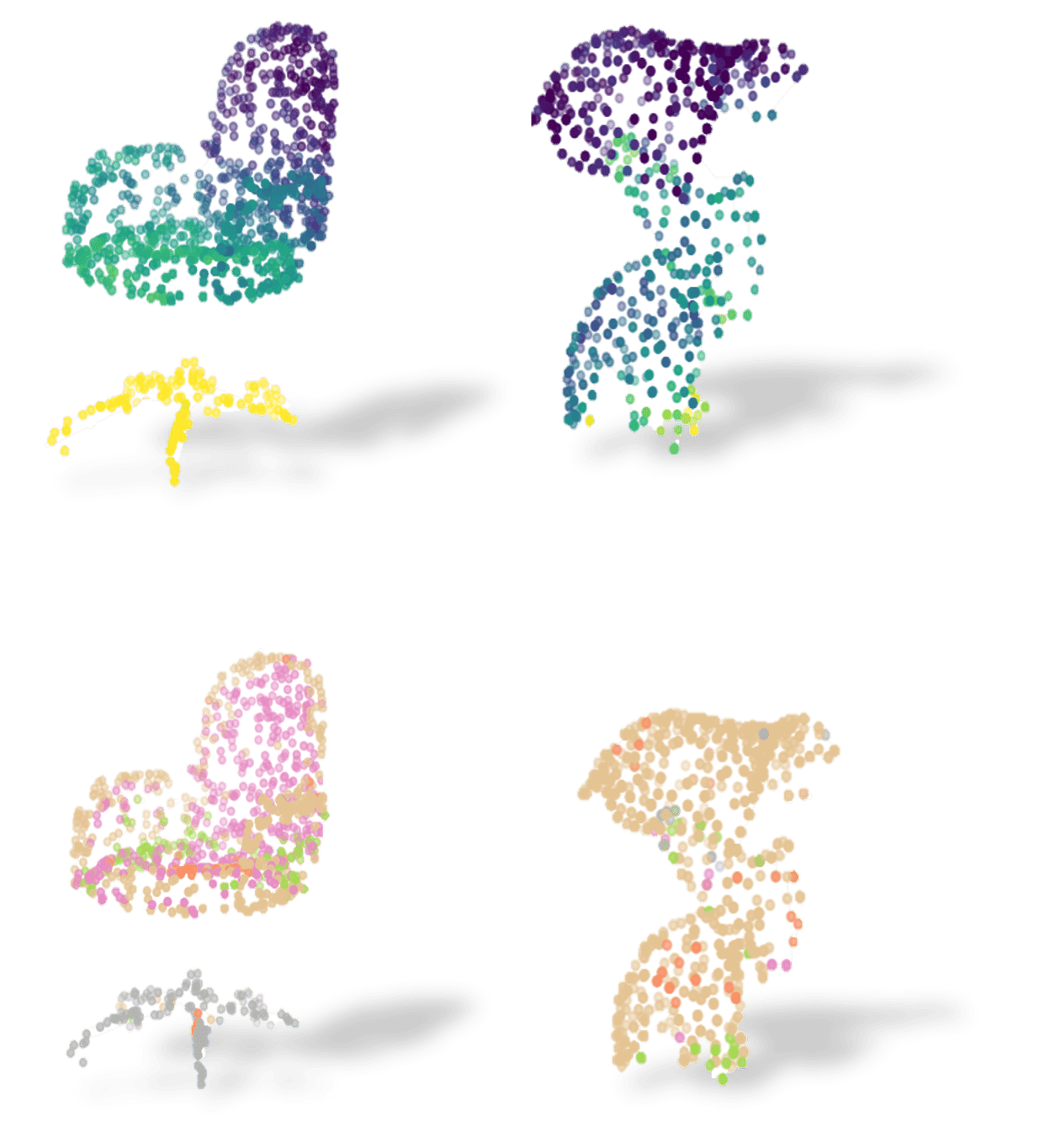}
        \caption{ScanNet10}

    \end{subfigure}
\caption{Instance segmentation (up) and semantic segmentation (down) results of various datasets by the pretrained model on Primitive3D. Specifically, instance embedding has been transformed into a 1-d vector by Principal Component Analysis (PCA). }
\label{fig:feature_learning}   
\end{figure}

\textbf{Visualization of Pretraining Result.} 
We provide the visualizations in Figure \ref{fig:feature_learning} to show the benefits of Primitive3D annotations on the object understanding. As it shows, the primitive segmentation task may aid deep learning models in decomposing objects.  Thus, it can serve as a low-level task, reducing the effort required for downstream semantic learning. However, certain unstructured and noisy objects in the real-world dataset, such as \textit{lamp} in the last column of Figure \ref{fig:feature_learning}, might cause the pretrained model to fail.

\textbf{Study of Dataset Generation.}
We consider how the RCT structure and size of the Primitive3D pretraining dataset influence the benchmark classification result. The experiment setting is the same as the cross dataset evaluation in Section~\ref{subsection:5.2} but with different Primitive3D datasets.

\begin{figure}[h]
\centering
\begin{minipage}[t]{0.48\linewidth}
\centering
\includegraphics[width=1.05\linewidth]{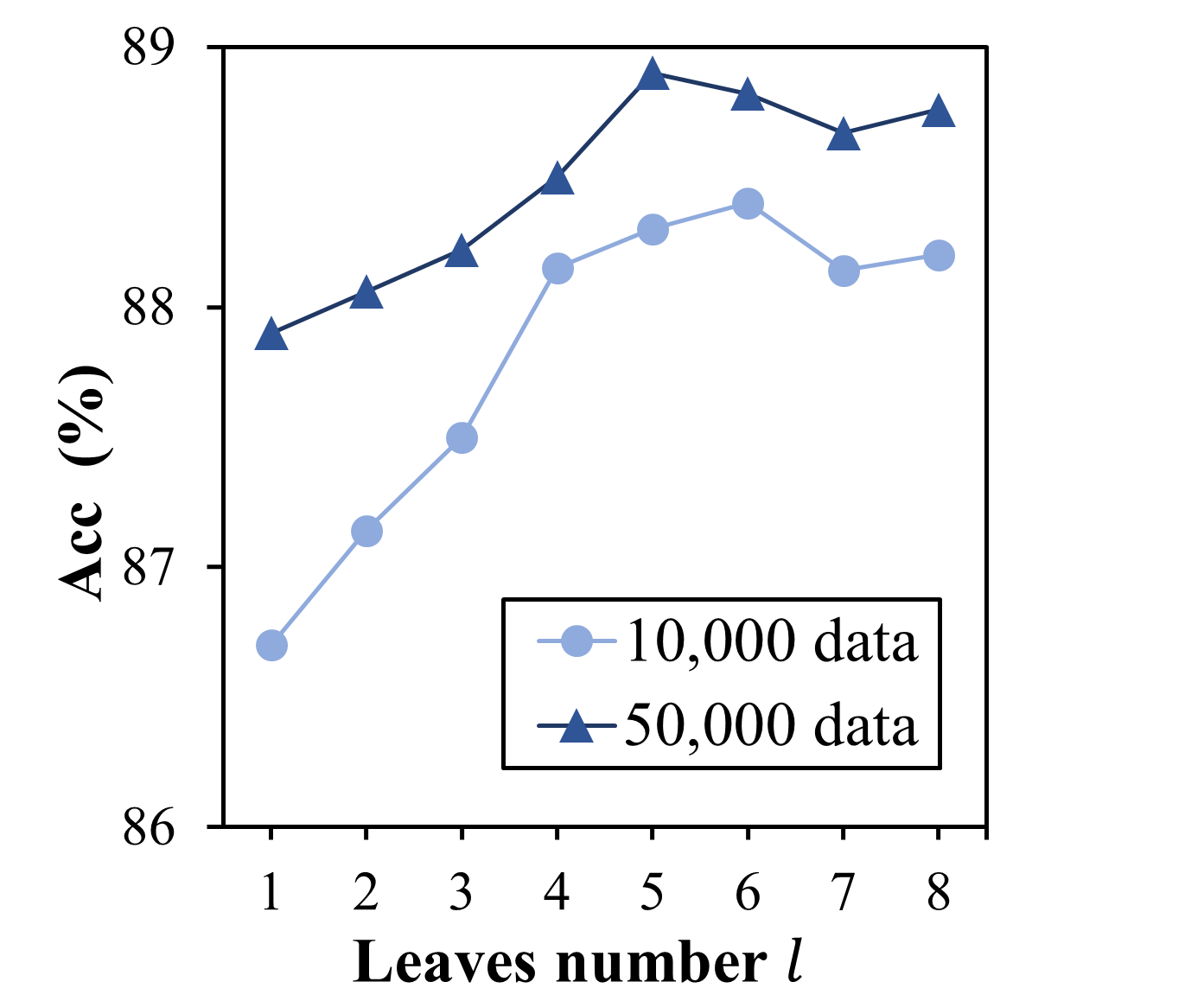}

\end{minipage}
\begin{minipage}[t]{0.48\linewidth}
\centering
\includegraphics[width=1.05\linewidth]{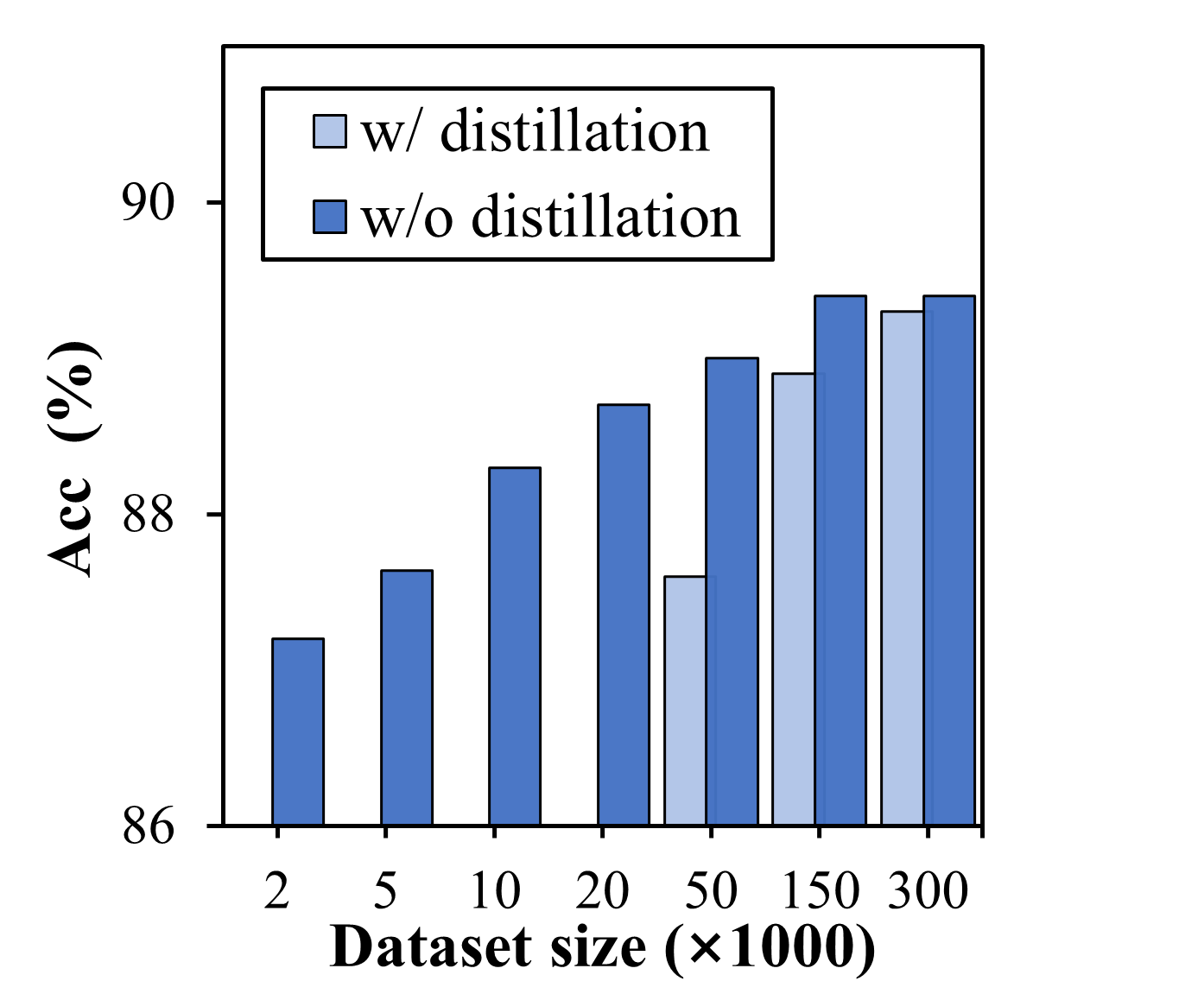}
\end{minipage}
\vspace{-0.5em}
\caption{Effect of leaves number (left) and effect of dataset size (right) of Primitive3D on ModelNet40 classification accuracy.}
\label{fig:dataset_effect}
\vspace{-0.5em}
\end{figure}

In Figure~\ref{fig:dataset_effect} (left), we depict how the accuracy changes as the number of RCT leaves $l$ varies. It indicates that the accuracy gain of increasing $l$ would get saturated once $l$ exceeds 6.
On the other hand, the time to generate an RCT sample rises considerably as $l$ increases due to the growing complexity in the execution of the boolean operations.
To strike a balance between performance and time, we limit $l$ of RCTs in our implementation to be within $[1,6]$. More time profiles of data generation can be found in Appendix. 

In Figure~\ref{fig:dataset_effect} (right), we generate Primitive3D datasets with sizes ranging from 2,000 to 300,000 and same RCT settings, and perform the benchmark classification. The result suggests that increasing the size of the dataset, whether dataset distillation is applied or not, improves the accuracy; however, the performance gain of increasing size is minor once it goes beyond 150,000. For this reason, we set the dataset size to be 150,000 in our experiments.

\vspace{-0.5em}

\begin{figure}[h]
\centering
\begin{minipage}[t]{0.47\linewidth}
\centering
\includegraphics[width=1.05\linewidth]{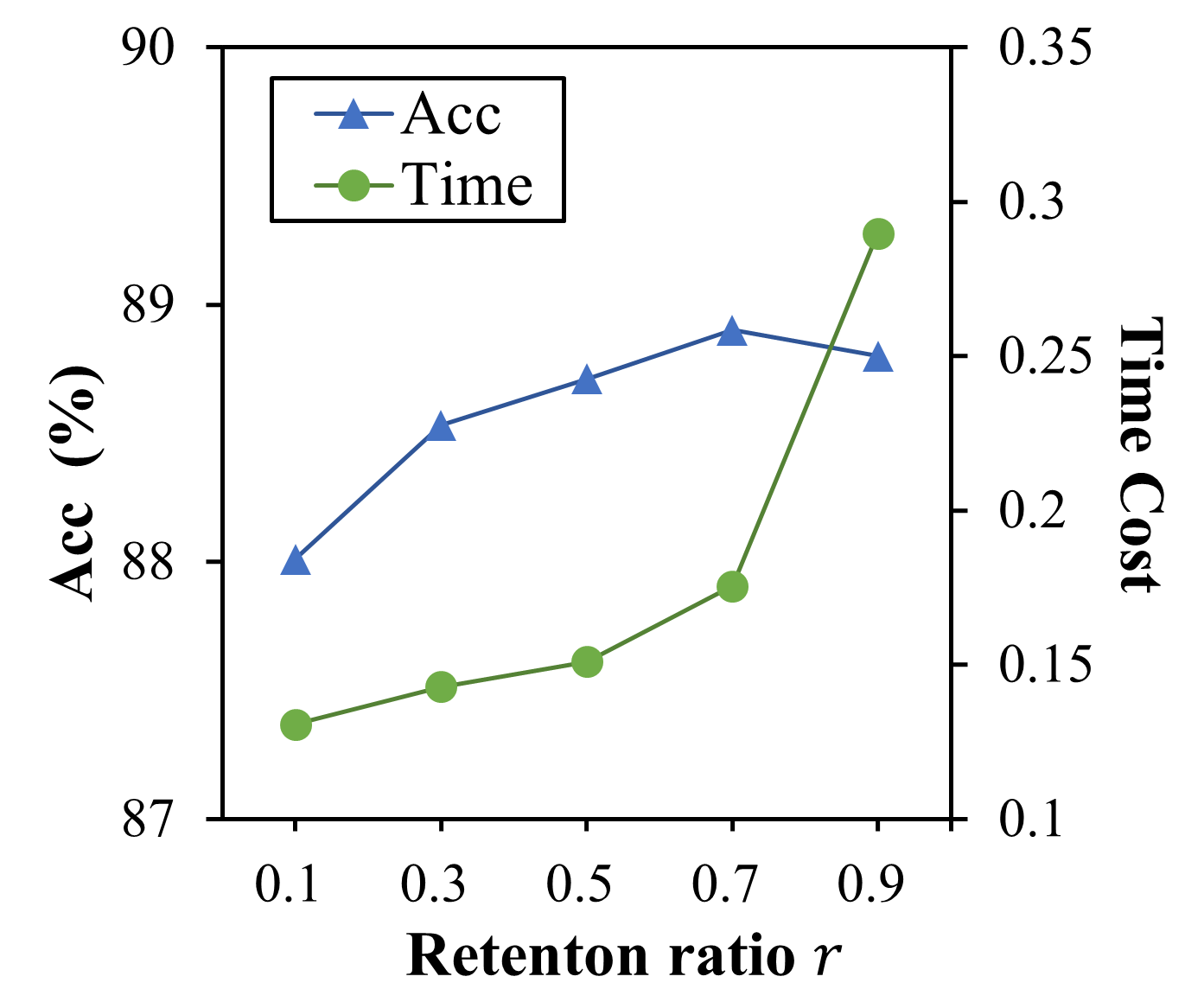}
\end{minipage}
\begin{minipage}[t]{0.47\linewidth}
\centering
\includegraphics[width=1.05\linewidth]{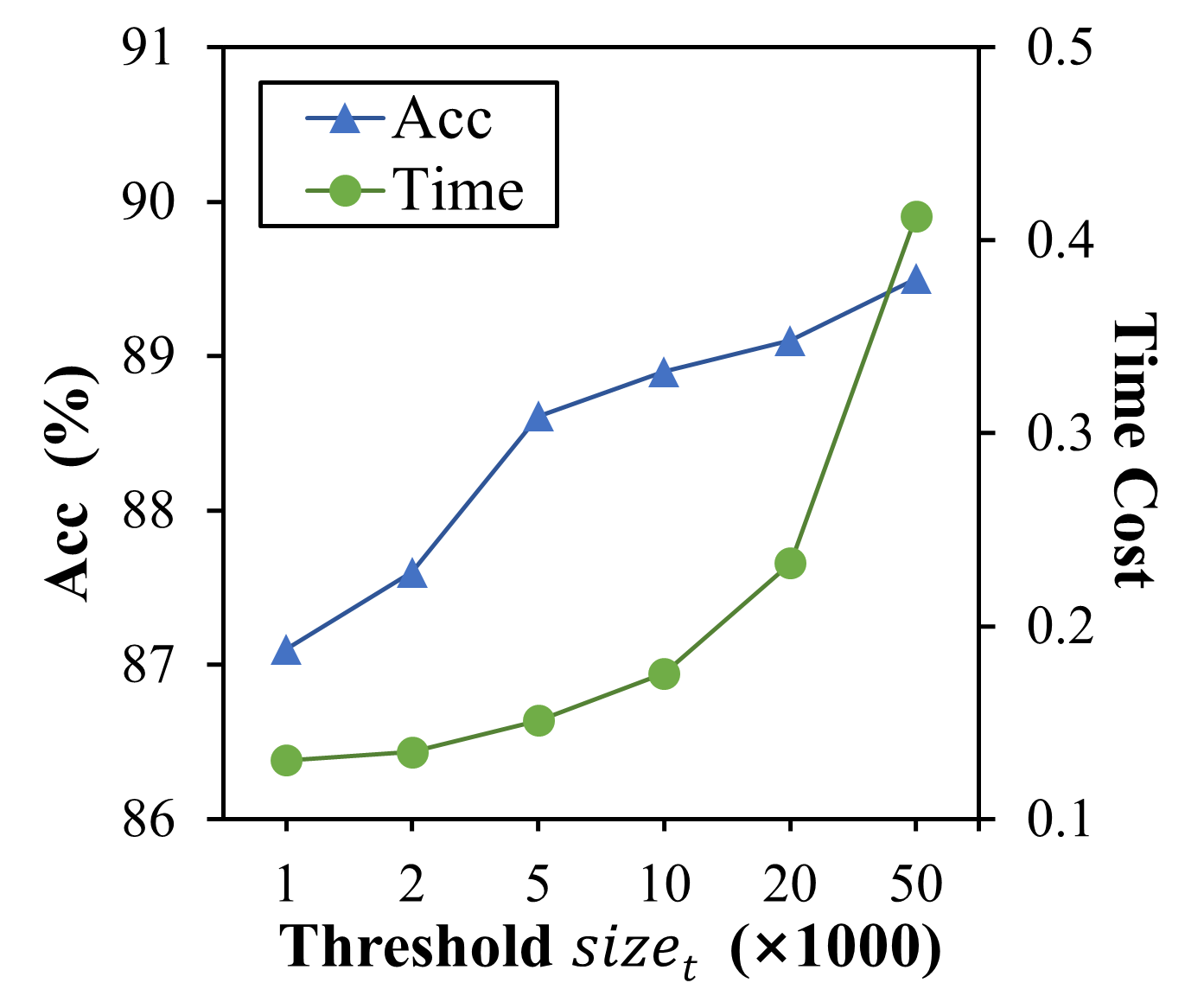}
\end{minipage}
\vspace{-0.8em}
\caption{Effect of retention rate (left) and threshold (right) in dataset distillation on the results of ModelNet40 classification.}
\label{fig:distill_effect}
\end{figure}

\textbf{Study of Dataset Distillation.} We perform ablation studies on how to choose the parameters for dataset distillation, \ie, the retention ratio $r$ and the dataset size threshold $size_t$. The experiment setting is the same as the cross dataset evaluation in Section~\ref{subsection:5.2} but with the incorporation of dataset distillation.
In Figure~\ref{fig:distill_effect}, we see that both the pretraining time and classification accuracy grow as the increases of $r$ and $size_t$, while the growth rate against $r$ is more subtle than the rate of $size_t$. 
Notably, when $size_t$ is set to 1,000, the accuracy drops to 87\% due to the excessive dropping of pretraining data, yet the consumption time remains more than 0.13.  It is also seen that a $size_t$ of 50,000 can reach a performance comparable to full Primitive3D. 
In our implementation, we let $r=0.7$ and $size_t=10,000$ to achieve a compromise between downstream task performance and time efficiency.  

\begin{figure}[h]
\footnotesize
\setlength\tabcolsep{-0.3pt}{
\begin{tabular}{c|cc|cc}
Target data & \multicolumn{2}{|c}{Retained samples} & \multicolumn{2}{|c}{Removed samples} \\
\begin{minipage}[b]{0.2\columnwidth}
	\centering
	\raisebox{-.4\height}{\includegraphics[width=1.0\linewidth]{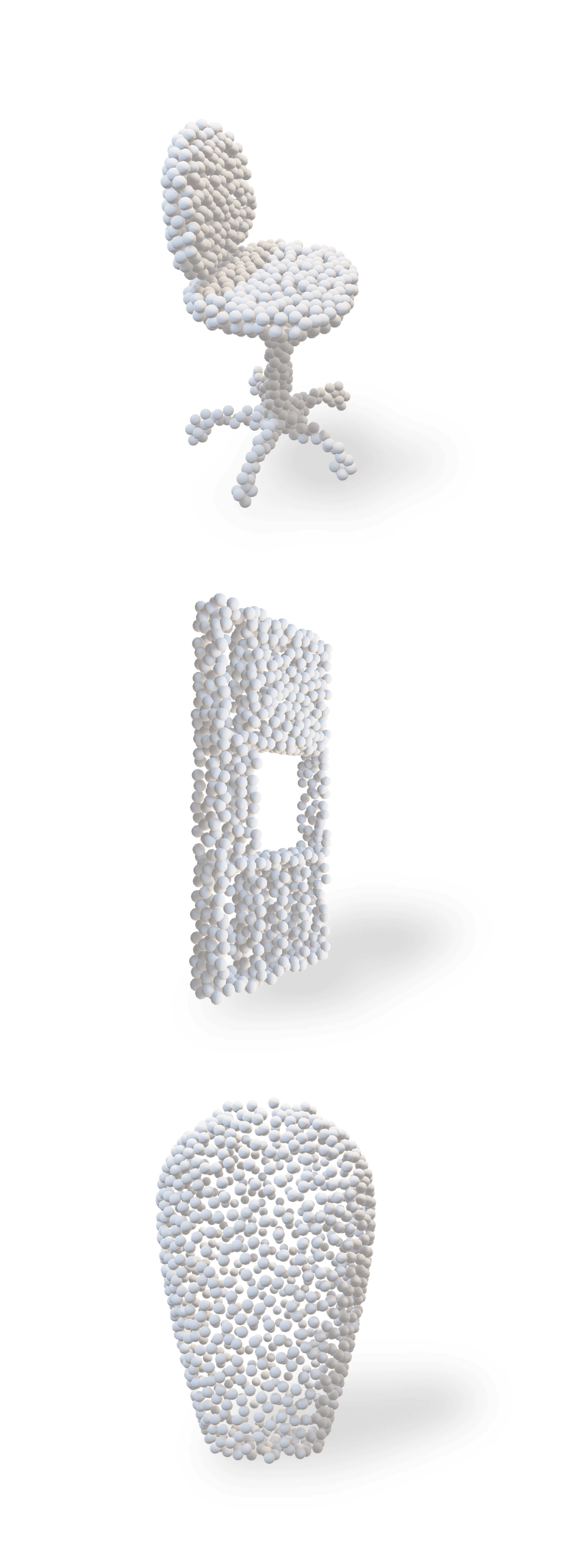}}
\end{minipage}               
           
&\begin{minipage}[b]{0.2\columnwidth}
	\centering
	\raisebox{-.4\height}{\includegraphics[width=1.0\linewidth]{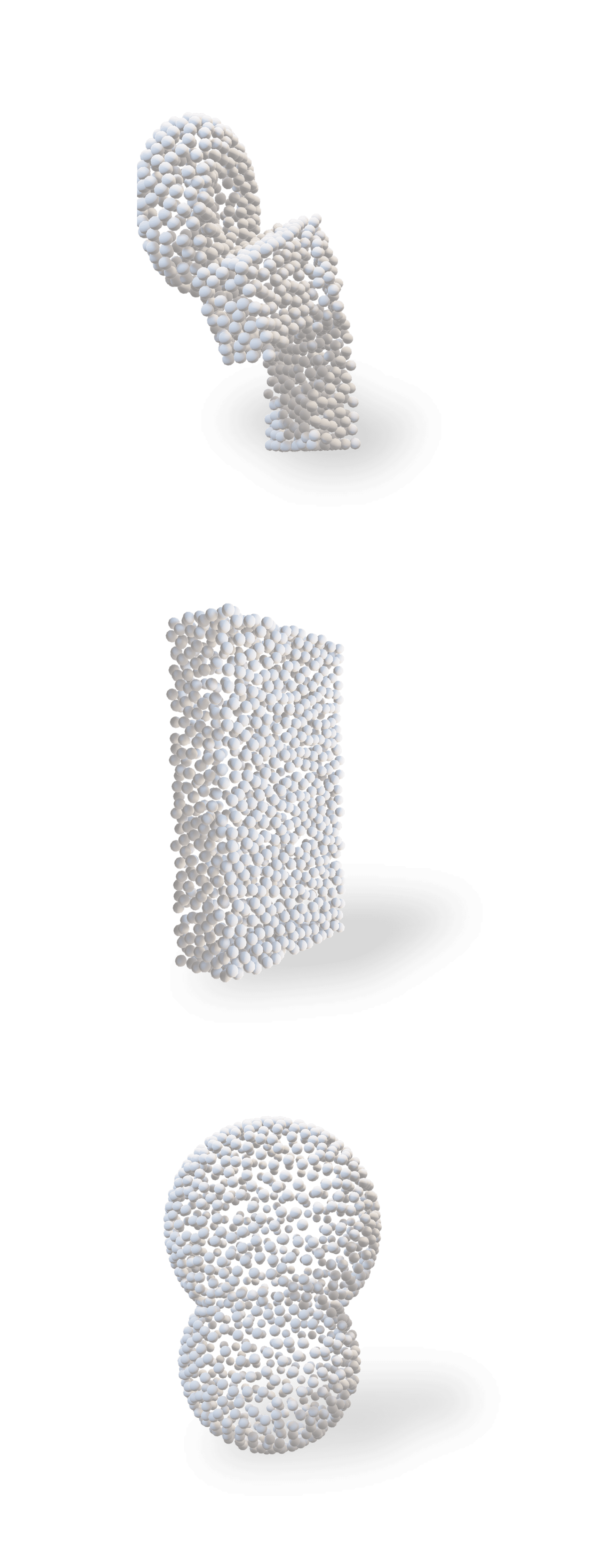}}
\end{minipage}                
&\begin{minipage}[b]{0.2\columnwidth}
	\centering
	\raisebox{-.4\height}{\includegraphics[width=1.0\linewidth]{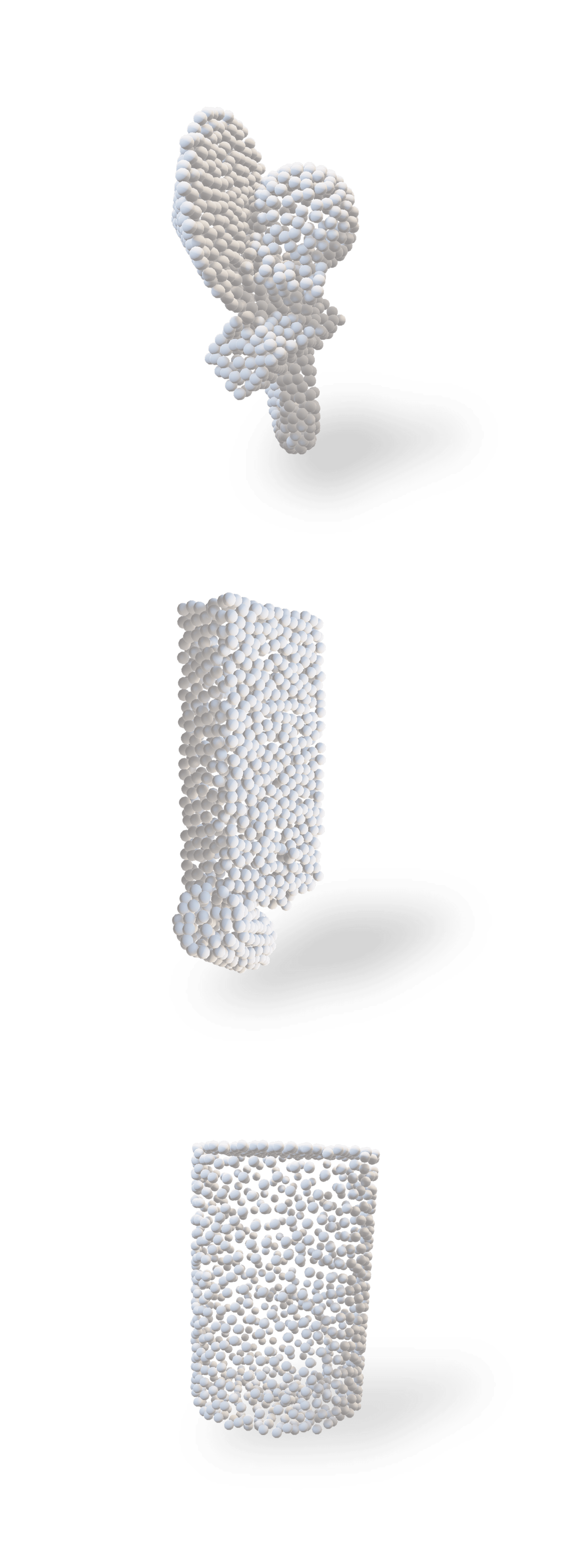}}
\end{minipage}                
&\begin{minipage}[b]{0.2\columnwidth}
	\centering
	\raisebox{-.4\height}{\includegraphics[width=1.0\linewidth]{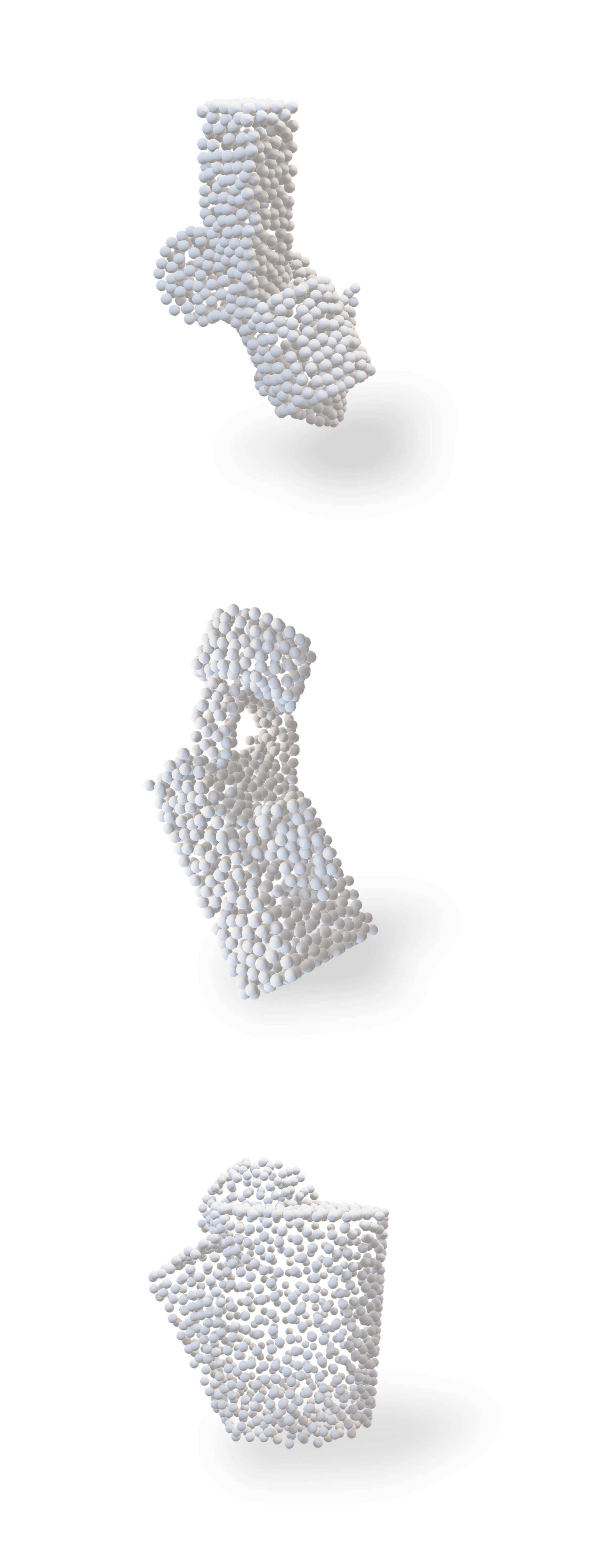}}
\end{minipage}                  
&\begin{minipage}[b]{0.2\columnwidth}
	\centering
	\raisebox{-.4\height}{\includegraphics[width=1.0\linewidth]{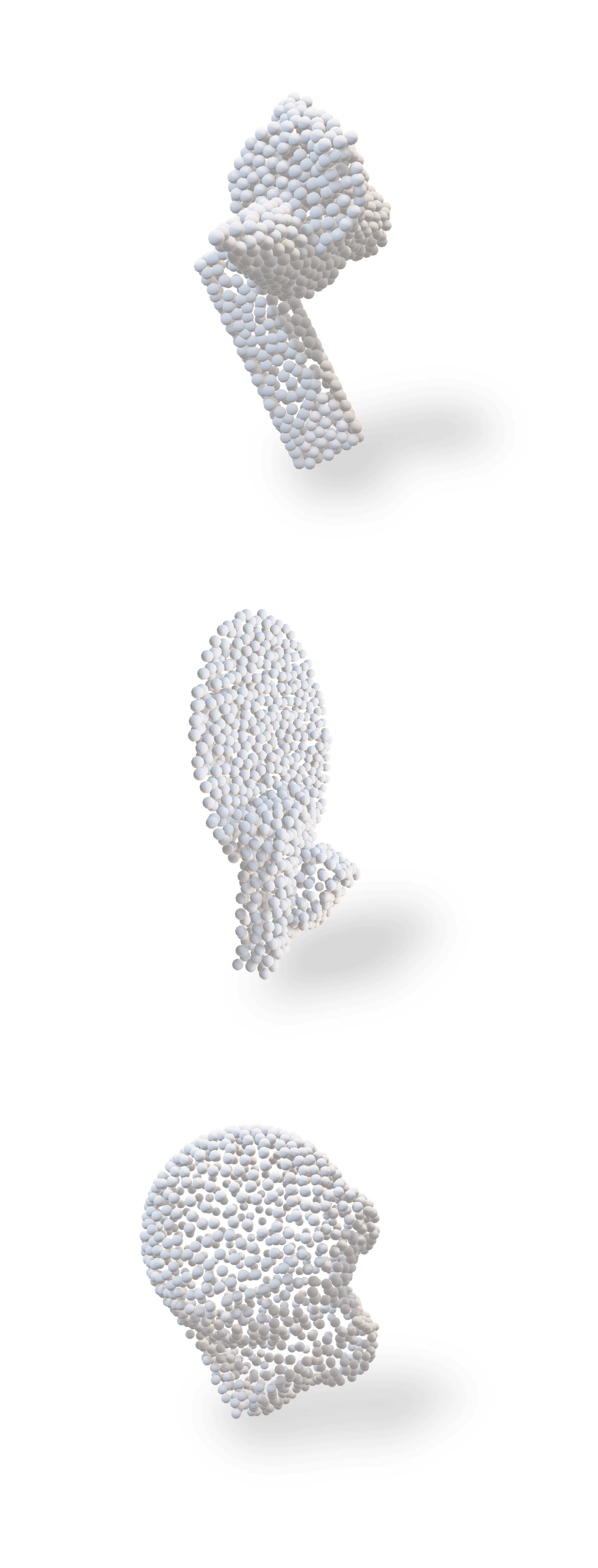}}
\end{minipage}                  \\
           
\end{tabular}}
\vspace{-0.12in}
\caption{Visualization of dataset distillation.}
\label{tab:vis_distillation}
\end{figure}

To visualize the dataset distillation, Figure~\ref{tab:vis_distillation} displays some retained and removed samples in the first dataset distillation stage. Particularly, the third column includes the nearest neighbors of the target data in the feature space from the removed samples. As can be observed, data distillation removes samples from the Primitive3D set that are not similar to the target data, so as to reduce the cost of learning irrelevant data. 

\textbf{Study of Pretraining Tasks.}
In Table~\ref{tab:ablation_task}, we study the effect of different pretraining tasks on the quality of the output features. Particularly, we consider four tasks, namely, unsupervised reconstruction with the original Chamfer distance, unsupervised reconstruction with the augmented Chamfer distance,  supervised semantic segmentation and supervised instance segmentation. \begin{table}[h]
\centering
\small
\begin{tabular}{cccc|c}
\toprule
 $U^{CD}$                   &$U^{ACD}$         & $S^t$       & $S^i$                                 & Accuracy (\%) \\ \hline
\checkmark       &         &                           &                                                     &85.6     \\
                 &\checkmark        &                           &                                                      &86.1     \\
                 &     &                           & \checkmark                                                 &87.1 \\
                 &     &\checkmark                 &                                                                &88.5     \\
                 &     & \checkmark                & \checkmark                                                      &88.9     \\
                 &\checkmark     & \checkmark                            & \checkmark                                                       &\textbf{89.4}     \\ \bottomrule
\end{tabular}
\caption{Accuracy of ModelNet40 classification. $U^{CD}$ and $U^{ACD}$ represent the implementation of $\mathcal{L}_{inst}$ with the Chamfer distance and the augmented Chamfer distance, respectively. $S^t$ and $S^i$ stand for applying the tasks of $\mathcal{L}_{type}$ and $\mathcal{L}_{inst}$, respectively.  }\label{tab:ablation_task}
\end{table}

We also test their combinations to see the compositional effect. The results suggest that semantic segmentation is the most effective individual task among all, while combined tasks can further enhance learning performance.

\textbf{Other Downstream Tasks \& Comparisons.}
Finally,  we investigate more downstream tasks and comparisons to other pretraining methods. We perform object part segmentation and unaligned object classification tasks. The former is a fine-grained shape recognition task based on ShapNetPart~\cite{yi2016scalable} dataset, on which our pretraining technique enables DGCNN to improve the mean IoU from 85.0\% to 85.3\%. 
Unaligned object classification is performed on the unaligned ModelNet40 dataset, \ie, randomly rotated objects. The feature representations obtained by our method outperform the supervised learned features by a large margin. 
Moreover, by fixing the pretraining dataset, we compare our technique with other SOTA pretraining methods. Experiments show that the highest performance is obtained when our strategy is used with Primitive3D. Note that Appendix C details the aforementioned experiments. 

\section{Conclusion}
We propose an efficient approach to generate 3D objects with part-based annotations based on a randomized construction process. The efficiency  allow us to synthesize a large-scale and densely-annotated 3D object dataset. To take advantage of it, we introduce a learning process that comprises multiple tasks and, optionally, a dataset distillation strategy. Combined with our generated dataset, the suggested learning produces well-generalized representations of 3D objects. 
The result of the experiments indicates that our dataset allows for better pretraining of models than other commonly used datasets. We expect our attempt to provide a new data source for training 3D deep models.

\section*{Acknowledgment}
This research is supported by the NUS Research Scholarship. The authors also sincerely thank Yongxing Dai and Chenxi Ma for their contributions to the paper.

{\small
\balance
\bibliographystyle{ieee_fullname}
\bibliography{egbib}
}

\end{document}